\documentstyle[aaai]{article}
\newcommand{\term}[1]{{\sc #1}}
\title{The Role of the Gricean Maxims\\
in the Generation of Referring Expressions}
\author{Robert Dale\\Microsoft Institute\\65 Epping Road\\North Ryde\\
Sydney {\sc nsw
2113}\\Australia\\{\tt rdale@mpce.mq.edu.au}\\
\And Ehud Reiter\\Department of Computer Science\\King's College\\
University of
Aberdeen\\Aberdeen {\sc ab9 2ue}\\Scotland\\{\tt ereiter@csd.abdn.ac.uk}} 
\begin{document}
\maketitle
\begin{abstract}
Grice's maxims of conversation [Grice 1975] 
are framed as directives to be followed
by a speaker of the language.
This paper argues that, when considered from the point of view
of natural language generation, such a characterisation is rather misleading,
and that the desired behaviour falls out quite naturally if we
view language generation as a goal-oriented process.  We argue this
position with particular regard to the generation of referring
expressions.
\end{abstract}
\section{Introduction}
The position taken in this paper can be summarised
as follows.
\begin{enumerate}\setlength{\itemsep}{0in}
\item Grice's maxims [Grice 1975] are framed as directives to the speaker,
and so it is natural to consider how they might impact on the task of
natural language generation ({\sc nlg}).
\item A number of the maxims can collectively be expressed by the
imperative ``Don't say too much and don't say too little.''
This focusses our attention on the language generation subtask
of \term{content determination}; and one of the more constrained and
well-explored aspects of content determination is the generation of
referring expressions.  However, if we look at this task in detail,
it becomes clear that there are problems with enforcing
a literal interpretation of the maxims.
\item We review some of our previous work that has
tried to address this problem, but go on to suggest a rather more radical
position: that Grice's maxims are unnecessary directives from the
point of view of referring expression generation, and that, provided
the register and sublanguage conventions of the genre in force are
conformed to, the
behaviour the maxims characterise actually falls out quite naturally from
viewing {\sc nlg} as a goal-oriented process.

\end{enumerate}
Under this view,
the maxims are no more than {\it post hoc} characterisations of the way
language works, and their framing as directives is ultimately rather
misleading.

\section{Adequate and Efficient Referring
Expressions}\label{content-determination}

\subsection{Grice and Reference:  Deciding What To Say}

It has become commonplace to view language generation as encompassing
two kinds of concerns: deciding what to say, and deciding how to say
it.\footnote{We take no particular stance in the present discussion
as to how this distinction impacts on the modularity of 
the architecture of a language
generation system.}  
The considerations that arise in deciding what to say are often
expressed in terms that echo Grice's Maxims of Quantity: don't say too
much, and don't say too little.  There are many reasons why
such imperatives are worth attending to in the development of
algorithms
to be used by computational systems which generate natural
language; for example,
we want to make sure that we have
given the hearer the information that she needs, but we don't
want to bore her with a flood of unnecessary statements.
{}From the point of view of implicatures, however, an additional concern
is that saying too much might lead the hearer to read between the
lines in ways that were unintended by the underlying system.
Ultimately, language generation systems should be as capable of
{\em exploiting} the notion of conversational implicature as much as people
do; but before we can achieve goals of that kind, it's important that
we know when we are {\em obeying} the maxims.
In other words, our first priority is to ensure that the generated text
does not unintentionally contain false implicatures.

The task of generating
referring expressions---and in particular, anaphoric referring
expressions---provides an arena where we can move towards a more
formal specification of what this really involves.
Given some internal symbol that corresponds to an intended referent,
the job of a referring expression generator is
to determine the semantic content of a noun phrase that will
identify the intended referent to the hearer.
The first serious consideration of this issue in the {\sc nlg}
literature was probably McDonald's
[1980] discussion of {\term potential distractors}---other entities in the
context we might mistakenly refer to---when deciding
whether or not it was safe to use a pronoun to refer to an entity.
Appelt [1982] and Novak [1988] looked at determining the content of
definite noun phrase referring expressions
in situations where (for whatever reason)
a pronoun could not be used.
In Dale [1989], one of
us characterised the task of determining the content of a referring
expression as being constrained by three Grice-like principles:
the \term{principle of sensitivity}, which states that the referring
expression chosen should take account of the state of the hearer's
knowledge; the \term{principle of adequacy}, which states that the referring
expression chosen should be sufficient to identify the intended
referent; and the \term{principle of efficiency},
which states that the referring
expression chosen should provide no more information than is necessary
for the identification of the intended referent.

\subsection{An Algorithm for Saying The Right Amount}

Suggestions that referring expression generation
should be governed by principles like those just described are common,
but detailed algorithms that meet the specified goals are somewhat rarer.
Dale [1989] proposed an algorithm which
assumes the following scenario:
\begin{quote}
Suppose that we have a set of entities $C$ (called the \term{context
set}) such that $C = \{a_{1},a_{2},\ldots, a_{n}\}$; and our task is
to distinguish from this context set some intended referent $r$ where
$r \in C$.  Suppose, also, that each entity $a_{k}$ is described in
the system's knowledge base by means of a set of properties,
$p_{k_{1}}, p_{k_{2}}, \ldots, p_{k_{m}}$.  In order to distinguish
our intended referent $r$ from the other entities in $C$, we need to
find some set of properties which are together true of $r$, but of no
other entity in $C$.  The linguistic realisation of this set of
properties constitutes a \term{distinguishing description} ({\sc dd})
of $r$ with respect to the context $C$.  A \term{minimal
distinguishing description} is then the linguistic realisation of the
smallest such set of properties.
\end{quote}
The detail of the algorithm is unimportant for present purposes;
basically, it consists of three steps as follows:
\begin{enumerate}\setlength{\itemsep}{0in}
\item Check Success:  see if the description we have constructed so
far
picks out only one entity in the context.  If it does, stop.
If not, go to Step 2.
\item Choose Property:  determine which property of the intended
referent would rule out the largest number of other entities in the
context.  Go to Step 3.
\item Extend Description:  add the chosen property to the description
being constructed, and remove the entities it rules out from the context.
Go to Step 1 with this extended description and the reduced context.
\end{enumerate}
Reiter [1990] noted a serious deficiency of this algorithm:
it will not in fact always produce a minimal distinguishing description,
and indeed finding a minimal distinguishing description is equivalent
to solving a
minimal set-cover
problem, which is {\sc np}-hard.  The computational complexity of this task
raises the question of whether it is appropriate to insist on an algorithm
that creates minimal distinguishing descriptions. In other
words, it may be unreasonable to try to construct maximally adequate
and efficient referring expressions; or, more to the point, meeting
Grice's requirements of saying neither too much nor too little, if
taken literally, may be computationally unachievable.

\section{Cooperative Behaviour as an Epiphenomenon}\label{implicit-grice}

\subsection{Allowing in Redundancy}

One response to Reiter's objection is to take the view that the notion
of minimality sought in the algorithm above is too strong.  In
subsequent work we took a step back and asked: what do {\em people}
actually do when they construct referring expressions?  It is very
difficult to make any strong claims on the basis of the experiments
that have been done, but it does seem to be the case that people do
not build minimal distinguishing descriptions in the strong sense
suggested above.  We explored these
considerations in more detail in [Dale and Reiter 1995], where we
proposed a revised algorithm which is computationally efficient at the
cost of producing some informational redundancy in the referring
expressions it generates.

\subsection{Implicit vs Explicit Pursuance of the Maxims}

So:  obeying the Gricean maxims looks computationally problematic, and
it seems not to be what people do (with some caveats:
we are assuming a literal interpretation of the maxims, and
assuming that it is possible to generalise from
the experimental results).

Our early attempts to generate referring expressions (as presented,
for example, in 
Dale [1989] and Reiter [1990]) explicitly enforced variants of the
Gricean Maxim of Quantity.  However, our current hypothesis is that
this is in fact unnecessary.  We now take the view that it is a mistake
to view the Gricean Maxims as directives; they are really no more than
{\it post hoc} characterisations of what is going on.  They may even mislead
us in the construction of mechanisms that cooperate conversationally.

It is generally accepted that language generation can fruitfully be
viewed as a goal-oriented process.  In other words, a natural language
generation system may be given as input an agenda of goals 
that are to be satisfied in
the text being constructed; the system's 
task is to find linguistic devices that
satisfy each goal, removing the goal 
from the agenda once it has been achieved.
There are a wide variety of goals that might appear in such a
mechanism.  In the context of referring expression generation, typical
goals could include:
\begin{itemize}\setlength{\itemsep}{0in}
\item Getting the hearer to
identify the intended referent $r$.
\item Alerting the hearer to the fact that $r$ has the property
represented by the attribute value pair $\langle a, v\rangle$.
\end{itemize}
A goal of the form of the first of these will give rise to the
construction of a distinguishing description; a goal of the form of
the second of these might result in the inclusion of information
beyond that required for referent identification (Robin [1994] is a
good discussion of some of the issues that arise in developing
algorithms to achieve such goals).  This
information could be expressed in a separate clause, but could equally
well be folded into the same referring expression that is being used
for the referent identification in the first goal (Appelt [1982]
provides a very nice example of this).

The fact that information can appear in a noun phrase for purposes
other than referent identification means, of course, that the hearer
has to do some work in determining what the role of each provided
descriptor is.  In the case of an utterance like {\it Give me the red
pen}, the speaker may be providing the term {\it red} in order
to distinguish the intended referent from another pen which is green.
It is equally possible, though, that there is only one pen in the
context, and {\it red} is included in the description because colour has
special salience (it may be easier for the speaker to first look for red
objects, and then find the
particular red object which is a pen).
Another possibility---perhaps a little tenuous in the
current example, but clearly a possibility nonetheless---is that the
hearer may be red--green
colour blind, and the speaker is imparting additional
information about the colour of the pen
which the hearer may be able to make use of later.
A more common clue to descriptor purpose is that fact that
some properties are more likely to be used for referent identification
than others.
In the utterance {\it Sit
by the newly-painted table}, for example, the property {\it newly-painted}
could be being used to distinguish the intended referent from other
tables
in the context, but it is rather more likely that its purpose is to
warn the hearer not to put her elbows on the
table.\footnote{This example is due to Bonnie Webber.}

In addition to goal orientation, aspects of genre
such as register and sublanguage also
play an important role in determining appropriate referring expressions.
In particular, whether a specific referring expression is interpreted by the
hearer as being purely for identification or not may depend on the current
genre.  For example, in casual
conversation, a hearer might interpret {\it Give me the Staedtler pen}
as having some purpose beyond simple identification
(perhaps informing the hearer
that the speaker prefers pens made by Staedtler), since manufacturer is
not a commonly used attribute in identification-only referring expressions
in this genre.  In an inventory-stocking context, on the other hand,
{\it Give me the Staedtler pens} might
be construed as purely referential, since manufacturer is often used
as an identifying attribute in this genre.
Consequently, an {\sc nlg} system that is generating an
identification-only referring expression should
if possible use only those attributes that are typically
used for identification
in the target genre; otherwise, false implicatures may arise.
However, again we believe that there is no need to explicitly model this
phenomenon as an implicature; it is sufficient to design the system so
that it uses the identifying attributes preferred in its target genre
(as is done via the {\sc PreferredAttributes} list in the algorithm of
[Dale and Reiter 1995]).

Although in the above examples the hearer may have to perform some
potentially complex inferencing to determine what the speakers'
goals are, note that there is no need for the speaker to do anything other
than satisfy
the list of goals using resources appropriate to the current genre.
Nowhere is there an explicit attempt to adhere to the maxims.

\subsection{Reassessing the Maxims}

In the light of the above discussion, we revisit Grice's maxims in this
section and comment on how each might be best interpreted in the
context of natural language generation.

\subsection{The Maxim of Quality}

{\it Try to make your contribution one that is true. More specifically:
\begin{enumerate}\setlength{\itemsep}{0in}
\item Do not say what you believe to be false.
\item Do not say that for which you lack adequate evidence.
\end{enumerate}}
\noindent No natural language generation systems that we are aware of
deliberately say things that are false:  this can happen by accident, of
course, but then it is not intentional.

An arguable exception
to this claim is the work of Jameson [1987], whose Imp system injects
bias into its utterances in order to mislead; but even here this is
not done by telling lies.  Certainly, in
principle one could construct a generation system that `lied' for
purposes such as advertising or manipulation, 
or that produced descriptions that were `correct'
relative to the hearer's knowledge even if they were untrue in the world;
for example, we might want to build
a system which could generate {\em the man drinking a martini}
to refer to a man who was actually drinking water from a martini glass.
This would require explicit programming, however; the default behaviour
of all systems we are aware of is to automatically obey the Maxim of Quality.

\subsection{The Maxim of Quantity}

{\it \begin{enumerate}\setlength{\itemsep}{0in}
\item Make your contribution as informative as is required (for the
current purposes of the exchange).
\item Do not make your contribution more informative than is required.
\end{enumerate}}
\noindent The first part of the Maxim of Quantity is automatically
fulfilled by a goal-oriented system: the goal will not be satisfied
until sufficient information is provided.  What we should say about
the second part of the maxim, however, depends on how strongly or
literally we choose to interpret it.  If we insist that referring
expressions or other utterances contain no unnecessary words, then we
will probably have to explicitly enforce this as a constraint in our
{\sc nlg} system; in general, {\sc nlg} systems will not automatically
obey this rule.  On the other hand, if we interpret the second part of
the Maxim of Quality as meaning `do not go out of your way to add extra
information that is not needed', then this behaviour once more
comes for free
with goal-orientation.  Our experience suggests that, at least for the
task of generating referring expressions, the second interpretation is
the best one.

\subsection{The Maxim of Relevance}

{\it Be relevant.  }

\noindent Yet again, this follows directly from goal-oriented
behaviour:  there is no reason why the system should consider saying
something that is not relevant.  It is possible, of course, that an
algorithm might unintentionally include irrelevant information; this
is also true of human linguistic behaviour.

\subsection{The Maxim of Manner}

{\it Be perspicuous.  More specifically:
\begin{enumerate}\setlength{\itemsep}{0in}
\item Avoid obscurity of expression.
\item Avoid ambiguity.
\item Be brief (avoid unnecessary prolixy).
\item Be orderly.
\end{enumerate}}
\noindent These are places where an anticipation feedback mechanism
of the kind
proposed by Jameson and Wahlster [1982] might be appropriate:  i.e, we might
like a system to subject its proposed utterance to a self-monitoring stage,
to make sure that it is not ambiguous and so on.  Of all the maxims,
it is perhaps these (with the exception of the brevity submaxim,
to which our response is the same as our response to
the Maxim of Quantity) which are most amenable to explicit modelling
in the generation process; but
even here it is equally possible that over time we
learn heuristics that do the job for us, so that the generation task
more or less naturally produces results that have the required
characteristics (see Levelt's [1989] comments on this as a possible
characteristic of the human language production mechanism).

\subsection{Exploiting Violations}

The goal-oriented approach has the additional benefit
that using violations of the maxims in order to get some other point
across falls out as part of the same mechanism; again, see Jameson's
[1987] work in this regard.  From the point of
view of the generator they are not violations at all.

\section{Conclusions}\label{conclusions}

We have argued that Grice's Maxims do not need to be explicitly enforced
or modelled in a natural language generation system.  
Instead, they should be replaced
by the following system construction principles.

\subsection{Grice and Generation}

A generation system should be goal-driven, and conform to the current genre.
As a general architecture, this suggests a process which builds an agenda of
goals, and then searches for communicative and linguistics resources in the
target genre which can be used to realize these goals.
There is the possibility that such a system may
end up saying something
beyond what was intended.  That is acceptable.  Minimality is not
necessary; provided the information that is provided is there
because it serves some purpose, hearers will not make inappropriate
inferences.

\subsection{Grice and Interpretation}

As a corollary to the goal-oriented view of generation,
the hearer should assume that every
informational element in the speaker's utterance is there with some
intended purpose.  The hearer's job is then to work out what the speaker's
intended purpose is.  If we are in a context, for example, where
referent identification is obviously the task, and the expression
contains information unnecessary for identification, the hearer must
consider the possibility that 
this information has been provided by the speaker
for some other purpose---but
note that it might not be; it might have been put in to help with
identification even if it turns out that the hearer did not make use
of it for that purpose.\footnote{Returning to our earlier {\it red pen}
example, suppose the hearer just happens to be focussed on the red pen
when the utterance is produced.  Then, the property of redness
may have been included by the speaker in order to help the hearer
identify the intended referent even if it is not used for this purpose.}
Properties which are clearly not
able to help us in identifying the intended referent must be doing
something else.  In the context of our {\it newly-painted table}
example, it many cases it will be impossible to determine that
something is newly-painted simply by looking; and it is unlikely that
the speaker intends us to go around actually touching all the tables
to identify which one has that property; so it is reasonable to assume
that
the property has been provided for some other purpose.

\subsection{In Summary}

Ultimately, for the generator, Grice's maxims taken collectively mean
{\it Don't include elements that don't do anything.}
Our position is that, under a goal-oriented view of language
generation, there is no need to explicitly follow such a directive
at all; the
desired behaviour just falls out of the mechanism.  We have argued, in
the present paper, that this is true of the referring expression
generation task; it remains to be seen whether the same story can be
told of all language generation, and what the impact of this is on
models of language understanding.

\section*{References}

\begin{description}
\item[{\rm Douglas Appelt [1982{]}}]
Planning Natural Language Utterances to Satisfy Multiple Goals.
Technical Note 259, SRI International, Menlo Park, California.
\item[{\rm Robert Dale [1989{]}}]
Cooking Up Referring Expressions.
In {\it Proceedings of 27th
Annual Meeting of the
Association for Computational Linguistics}, Vancouver {\sc bc}.
\item[{\rm Robert Dale and Ehud Reiter [1995{]}}]
Computational Interpretations of the Gricean Maxims in the Generation
of Referring Expressions.
{\it Cognitive Science}, 19(2), pp233--263.
\item[{\rm H Paul Grice [1975{]}}]
Logic and conversation.
In P Cole and J Morgan, editors, {\em Syntax and Semantics: Vol 3,
Speech Acts}, pages 43--58. New York:  Academic Press.
\item[{\rm Anthony Jameson [1987{]}}]
How to Appear to be Conforming to the Maxims Even if you Prefer to
Violate
Them.  Chapter 2 in G Kempen (ed), {\it Natural Language Generation},
Dordrecht:  Martinis Nijhoff Publishers.
\item[{\rm Anthony Jameson and Wolfgang Wahlster [1982{]}}]
User Modelling in Anaphora Generation:  Ellipsis and Definite Description.
In {\it Proceedings of the Fifth European Conference on
Artificial Intelligence}, Pisa, Italy.
\item[{\rm Willem Levelt [1989{]}}]
Speaking:  From Intention to Articulation.
Cambridge, {\sc ma}: {\sc mit} Press.
\item[{\rm David McDonald [1980{]}}]
Natural Language Generation as a Process of Decision Making
Under Constraints.
PhD Thesis, {\sc mit}.
\item[{\rm Hans-Joachim Novak [1988{]}}]
Generating Referring Phrases in a Dynamic Environment.
Chapter 5 in M Zock and G Sabah (eds), {\it Advances in Natural
Language Generation}, Volume 2, pp76--85.  Pinter Publishers.
\item[{\rm Ehud Reiter [1990{]}}]
Generating Appropriate Natural Language Object Descriptions.
PhD Thesis, Harvard University.
\item[{\rm Jacques Robin [1994{]}}]
Revision-Based Generation of Natural Language Summaries
Providing Historical Background.  PhD Thesis, Columbia University.
\end{description}
\end{document}